%% file: main.tex
\documentclass[10pt,twocolumn,letterpaper]{article}

\usepackage[pagenumbers]{cvpr} % To force page numbers, e.g. for an arXiv version

\input{preamble}
\usepackage[table]{xcolor}
\usepackage{colortbl}
\usepackage{multirow}
\usepackage{array}
\usepackage{makecell}
\usepackage{booktabs}
\usepackage{adjustbox}
\usepackage{pifont}
\usepackage{float}
\usepackage[hang,flushmargin]{footmisc}

\definecolor{cvprblue}{rgb}{0.21,0.49,0.74}
\definecolor{darkgreen}{RGB}{0,100,0}
\usepackage[pagebackref,breaklinks,colorlinks,allcolors=cvprblue]{hyperref}

\def\paperID{6537} % *** Enter the Paper ID here
\def\confName{CVPR}
\def\confYear{2026}

\title{VersaViT: Enhancing MLLM Vision Backbones via Task-Guided Optimization}

\author{Yikun Liu$^{1,2*}$, Yuan Liu$^{3\dagger}$, Shangzhe Di$^{1,2}$, Haicheng Wang$^{1,2}$, Zhongyin Zhao$^3$,\\[2pt] Le Tian$^{3}$, Xiao Zhou$^{3}$, Jie Zhou$^{3}$, Jiangchao Yao$^{2}$, Yanfeng Wang$^{1\dagger}$, Weidi Xie$^{1}$ \\[3pt]
$^{1}$School of Artificial Intelligence, Shanghai Jiao Tong University, China \hspace{0.5cm} \\
$^{2}$CMIC, Shanghai Jiao Tong University, China \hspace{0.3cm}
$^{3}$ WeChat AI, Tencent Inc., China \hspace{0.5cm}
}

\begin{document}
\maketitle

 \renewcommand{\thefootnote}{\fnsymbol{footnote}} %将脚注符号设置为fnsymbol类型，即特殊符号表示
\footnotetext[1]{Work was done during internship in WeChat.}  
\footnotetext[2]{Corresponding author.} 

\input{sec/0_abstract}    
\input{sec/1_intro}

\input{sec/2_related-work}
\input{sec/3_motivation}

\input{sec/4_method}
\input{sec/5_experiments}
\input{sec/6_conclusion}

{
    \small
    \bibliographystyle{ieeenat_fullname}
    \bibliography{main}
}

\input{sec/X_suppl}


\end{document}

%% file: sec/0_abstract.tex
\begin{abstract}

% With the rapid advancement of Multimodal Large Language Models~(MLLMs), mainstream methods increasingly focus on jointly optimizing vision encoders and Large Language Models~(LLMs) solely through autoregressive loss, which demonstrates exceptional performance across a wide range of Visual Question Answering~(VQA) benchmarks. In this paper, we make the following contributions: (i) We first reveal that the vision encoders trained in this manner lose some vision-centric representation, leading to suboptimal performance on certain dense tasks (e.g., semantic segmentation, depth estimation). (ii) We propose WeViT, a \underline{we}ll-rounded vision transformer that instantiates a novel multi-task collaborative training paradigm, guiding the optimization of the vision backbone through lightweight task heads, thereby enhancing VQA performance while simultaneously strengthening its vision-centric representation. (iii) Extensive experiments across various downstream tasks demonstrate the effectiveness of our method, significantly enhancing the comprehensive capabilities of the vision backbone. The code and model will be made publicly available. \kun{edit last}

Multimodal Large Language Models (MLLMs) have recently achieved remarkable success in visual-language understanding, demonstrating superior high-level semantic alignment within their vision encoders. An important question thus arises: Can these encoders serve as versatile vision backbones, capable of reliably performing classic vision-centric tasks as well? To address the question, we make the following contributions: (i) we identify that the vision encoders within MLLMs exhibit deficiencies in their dense feature representations, as evidenced by their suboptimal performance on dense prediction tasks ({\em e.g.}, semantic segmentation, depth estimation); (ii) we propose VersaViT, a well-rounded vision transformer that instantiates a novel multi-task framework for collaborative post-training. This framework facilitates the optimization of the vision backbone via lightweight task heads with multi-granularity supervision; (iii) extensive experiments across various downstream tasks demonstrate the effectiveness of our method, yielding a versatile vision backbone suited for both language-mediated reasoning and pixel-level understanding. The project page is available \href{https://code-kunkun.github.io/VersaViT/}{here}.

\end{abstract}

%% file: sec/1_intro.tex
\section{Introduction}
\label{sec:intro}

\begin{figure}[ht]
  \centering
  \includegraphics[width=.47\textwidth]{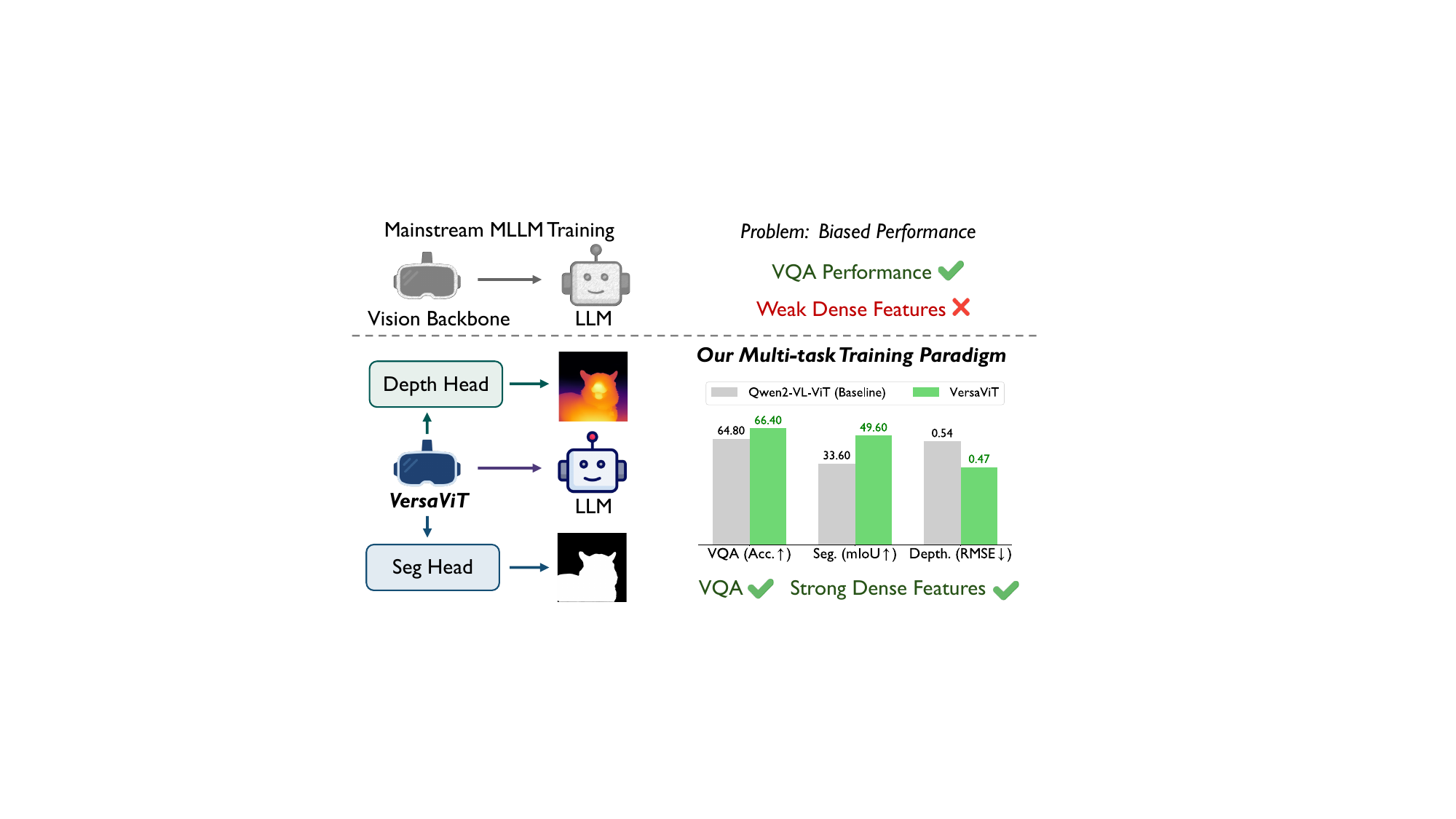} \\
  \vspace{-6pt}
  \caption{\textbf{Overcoming the dense feature limitation of vision backbone within MLLMs.} We observe the vision encoder within MLLMs (Top), which yields strong VQA but suboptimal dense features. Conversely, our multi-task collaborative post-training (Bottom) is designed to overcome this limitation by comprehensively enhancing the vision encoder's capabilities.}
  \vspace{-2em}
\end{figure}

Multimodal large language models~(MLLMs) have emerged as the prevailing paradigm for vision-language understanding~\cite{deitke2025molmo, bai2025qwen2, team2025kimi, guo2025seed1}. 
Architecturally, contemporary MLLMs consist of a vision encoder, a vision–language projection module, and a large language model (LLM). Among these, the vision encoder plays a pivotal role in linking visual perception with language reasoning. In practice, it is commonly initialized from CLIP-style encoders~\cite{radford2021learning, tschannen2025siglip} trained with contrastive learning~\cite{oord2018representation}, and then jointly optimized with the LLM via instruction tuning on large-scale visual question answering (VQA) and captioning corpora. This pipeline has yielded state-of-the-art results on a broad spectrum of visual–language understanding tasks.

% 第二段揭示我们观察到的现象
Despite the excellent performance on instruction following and open-ended VQA~\cite{liu2024points1, coreteam2025mimovltechnicalreport, bai2025qwen2, guo2025seed1}, it remains unclear whether MLLM vision encoders retain the pixel-level perception and spatial precision, that are often required for classic vision-centric problems, for example, segmentation, depth estimation. This question is central if we aim to evolve these encoders into true vision foundation models (VFMs) that are equally competent at language reasoning and pixel-level perception. To probe this, we conduct linear probing on monocular depth estimation and semantic segmentation using the vision backbones of the Qwen-VL series~\cite{bai2025qwen2, wang2024qwen2}. Our initial results (Section~\ref{sec:motivation}) reveal notable deficiencies on dense prediction benchmarks, suggesting a representational misalignment: pretraining and instruction tuning emphasize global semantic alignment and language grounding, but do not sufficiently regularize the fine-grained visual features critical for dense tasks.

This observation motivates our central question: 
\textit{can we adapt MLLM vision encoders so that they simultaneously excel at vision–language reasoning and vision-centric dense prediction, yielding a genuinely well-rounded vision foundation model ?} 
We posit that the gap arises less from architectural limits and more from the training curriculum and supervision granularity. Specifically, contrastive pretraining followed by text-centric instruction tuning provides limited multi-scale spatial supervision and sparse pixel-level signals, which can erode fine-grained dense representation.

To this end, we introduce \textbf{VersaViT}, a well-rounded vision transformer obtained by post-training the vision encoder of existing MLLMs with multi-granularity supervision and lightweight task-specific heads. 
Our framework couples the shared backbone with auxiliary heads to (i) expose the encoder to complementary signals spanning high-level semantics, mid-level spatial reasoning, and pixel-level perception, while (ii) isolating task-specific conflicts in the heads rather than in the shared representation. Concretely, we instantiate three representative supervision families: (1) VQA and image captioning for semantic grounding and instruction-following, (2) monocular depth estimation for 3D-aware spatial structure, and (3) referring image segmentation for localized, language-conditioned pixel precision. This framework jointly optimizes the backbone and all auxiliary heads with a simple multi-task objective, biasing the shared representation toward transferable, dense-friendly features without sacrificing language grounding.

We validate VersaViT across a wide range of downstream tasks, including VQA, semantic segmentation, referring segmentation, monocular depth estimation, and image–text retrieval. Empirically, VersaViT improves VQA performance while substantially strengthening dense prediction accuracy, indicating that multi-granularity signals can be synergistic rather than competitive when mediated by lightweight heads. Moreover, our framework is modular and aligns with MLLM training pipelines, enabling drop-in integration with minimal engineering overhead.

In summary, we make the following contribution: (i) diagnosis. We identify a consistent shortfall in dense representations of mainstream MLLM vision encoders, manifested by underperformance on depth and segmentation under linear probing; (ii) method. We propose VersaViT, a well-rounded vision transformer that instantiates a multi-task collaborative post-training framework. This framework augments an MLLM vision encoder with lightweight auxiliary heads for VQA/captioning, depth estimation, and referring segmentation, delivering multi-granularity supervision to the shared backbone; (iii) evidence. Extensive experiments show that VersaViT simultaneously advances VQA and dense prediction, yielding a more balanced, transferable vision backbone suited for both language-mediated reasoning and pixel-level understanding.

%% file: sec/2_related-work.tex
\section{Related Work}

\noindent \textbf{Vision Foundation Models.} Developing a robust foundation model for computer vision remains a fundamental challenge within the field. Various approaches have been adopted to advance the development of vision foundation models, which can be broadly categorized into two scalable learning paradigms: self-supervised and weakly-supervised learning. Recent self-supervised methods, including MAE~\cite{he2022masked}, iBOT~\cite{zhou2021ibot}, and the DINO series~\cite{caron2021emerging, oquab2023dinov2, simeoni2025dinov3}, mainly use large-scale curated image datasets to construct suitable proxy tasks for learning generalizable visual representations. In contrast, weakly-supervised learning leverages web-scale image-text pairs for training. This paradigm includes models such as CLIP~\cite{radford2021learning}, ALIGN~\cite{jia2021scaling}, and SigLIP~\cite{zhai2023sigmoid}, which utilize contrastive learning~\cite{oord2018representation}, as well as more recent models like CapPa~\cite{tschannen2023image}, LocCa~\cite{wan2024locca} and OpenVision2~\cite{liu2025openvision}, together with the Qwen-VL series~\cite{bai2025qwen2, wang2024qwen2} and other MLLMs~\cite{yang2025kwai, guo2025seed1, team2025kimi, yin2025sailvit}, which rely on autoregressive loss for training from scratch or for fine-tuning the vision encoder to achieve improved visual representations. In this paper, we focus on enhancing the representation capability of vision encoders within MLLMs through task-guided post-training optimization.

\vspace{2pt}
\noindent \textbf{Multimodal Large Language Models.} With recent advancements in LLMs~\cite{brown2020language, touvron2023llama, jiang2024mixtral}, increasing attention has been directed toward MLLMs, which aim to align visual and textual modalities via visual instruction tuning. Recent studies~\cite{guo2025seed1, team2025kimi, wang2024qwen2, bai2025qwen2, wang2025advancing} either seek to enhance the performance of MLLMs across various VQA benchmarks by improving data scaling and selection, architectural design, and training strategies, or leverage the powerful capabilities of MLLMs to repurpose them for other visual-language tasks~\cite{li2025universal, liu2025lamra, lai2024lisa, wang2025object}, such as multimodal retrieval, temporal grounding, and image referring segmentation. However, the intrinsic quality and characteristics of the visual features learned by the MLLM's vision encoder remain largely unexplored and unaddressed. Therefore, we aim to investigate this critical issue and further leverage these insights to enhance the vision encoder's representation.

\vspace{2pt}
\noindent \textbf{Multi-task Learning.} Multi-task learning~\cite{caruana1997multitask, zhuang2025argus, crawshaw2020multi, vandenhende2021multi, yan2025learning, yang2025soccermaster, di2026revisiting} enables models to generalize across diverse tasks, fostering the development of versatile systems that no longer rely on task-specific architectures. In natural language processing~(NLP), this paradigm has been successfully realized by framing a wide range of text-related tasks within a sequence-to-sequence formulation, which has facilitated the emergence of LLMs~\cite{brown2020language, touvron2023llama}. Inspired by this paradigm, the multimodal domain has also witnessed a surge of research on unified multimodal models~\cite{wang2021ufo, wu2024visionllm, yan2025task, Ranzinger_2024_CVPR, heinrich2025radiov2}. These approaches typically reformulate multimodal tasks, which span from image captioning to visual question answering, as sequence-to-sequence problems or leverage LLMs as bridges to connect different task heads. Motivated by these developments, we explore the joint training of multiple vision tasks to enhance the vision backbone.

\begin{figure}[ht]
  \centering
  \includegraphics[width=.47\textwidth]{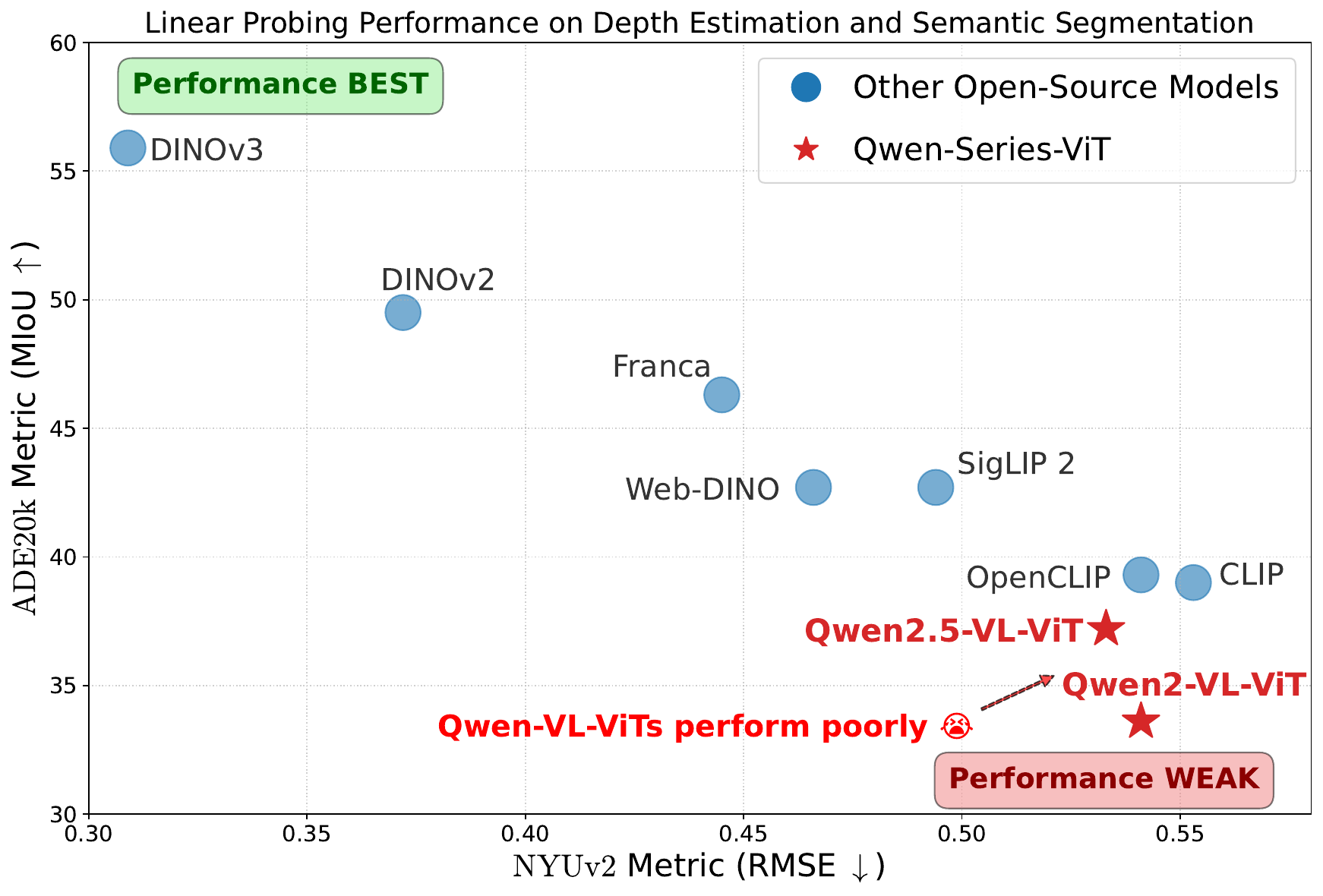} \\
  \vspace{-6pt}
  \caption{\textbf{Linear probing performance on depth estimation (NYUv2) and semantic segmentation (ADE20k) across different vision backbones.} The results reveal that the qwen-vl-vits exhibit suboptimal performance on these vision-centric benchmarks.}
  \vspace{-1em}
 \label{fig:motivation}
\end{figure}

%% file: sec/3_motivation.tex
\section{Motivation}
\label{sec:motivation}

% \weidi{this section now is a bit too long, try to make it more concise.}

Recent vision encoders developed within the MLLM paradigm have demonstrated exceptional proficiency in handling VQA tasks. However, a practical vision foundation model (VFM) should be broadly useful beyond dialog, supporting perception systems that require spatial precision, geometric consistency, and pixel-level fidelity. This raises a critical question: \textit{Are current MLLM vision encoders effective VFMs, or do they exhibit systematic gaps when applied to classic vision-centric tasks?} To explore this, we conduct a linear probing evaluation, assessing two representative encoders, Qwen2-VL-ViT~\cite{wang2024qwen2} and Qwen2.5-VL-ViT~\cite{bai2025qwen2}, on monocular depth estimation and semantic segmentation. As shown in Figure~\ref{fig:motivation}, both encoders underperform substantially in this setting. These results suggest a representational skew induced by the prevailing training recipe, which prioritizes global semantic alignment while under-regularizing intermediate and fine-grained visual structures.

% Why is this investigation necessary now? First, the field increasingly relies on MLLMs as default vision backbones, yet the community lacks principled evidence on their dense-transfer reliability. Second, large-scale end-to-end retraining is costly; if limitations stem from the supervision curriculum rather than architecture, a lightweight post-training path could be both effective and practical. Third, real-world applications, from robotics and AR to scientific imaging, demand encoders that are simultaneously language-aware and pixel-accurate. Therefore, determining whether such a balance is achievable with existing MLLMs is of immediate utility.

% Why is this investigation necessary now? First, the field increasingly relies on MLLMs as the default solution for various tasks, yet the community lacks principled evidence on their dense-transfer reliability. Second, large-scale end-to-end retraining is costly; if limitations stem from the supervision curriculum rather than architecture, a lightweight post-training path could be both effective and practical. Third, real-world applications—from robotics and AR to scientific imaging—demand encoders that are simultaneously language-aware and pixel-accurate. Establishing whether such a balance is achievable with existing MLLMs is therefore of immediate utility.

Guided by these findings, we ask: Can MLLM encoders be rebalanced toward dense-friendly features without sacrificing their language grounding via low-cost post-training? Our hypothesis is that multi-granularity supervision, combining high-level semantic grounding (VQA/captioning), mid-level spatial reasoning ({\em e.g.}, depth estimation), and pixel-level localization ({\em e.g.}, referring segmentation), can remedy the observed skew with minimal architectural change. We implement this idea in VersaViT, which is optimized via a collaborative multi-task post-training framework that attaches lightweight, task-specific heads to a shared vision backbone. The heads provide task-targeted gradients and absorb conflicts, while the vision backbone is nudged toward representations that are both instruction-competent and dense-friendly. This design intentionally prioritizes practicality: it is modular, compatible with standard MLLM pipelines, and avoids full-scale retraining.

In summary, our motivation is empirical and application-driven: current MLLM vision encoders fail to reliably apply to dense prediction tasks, despite excelling at VQA. We show that a low-cost, multi-granularity post-training can close this gap, yielding a more balanced and useful VFM.

%% file: sec/4_method.tex
\section{Method}

\begin{figure*}[ht]
  \centering
  \includegraphics[width=\textwidth]{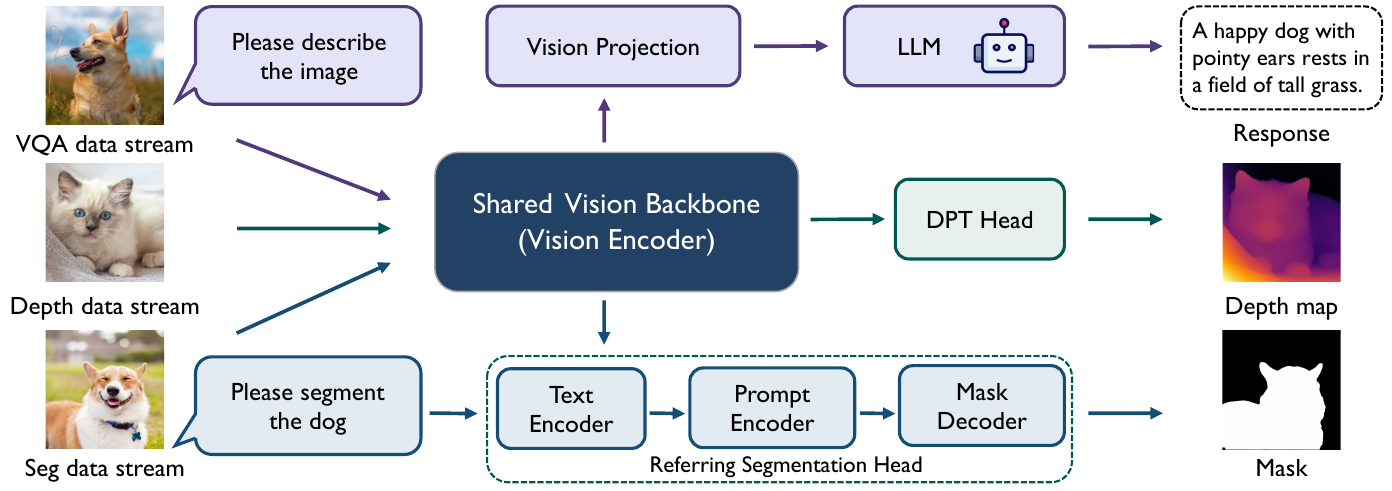} \\
  \vspace{-6pt}
  \caption{\textbf{Overview of the proposed multi-task collaborative training framework.} The proposed framework jointly trains three distinct tasks: VQA and Image Captioning, Monocular Depth Estimation, and Image Referring Segmentation. By incorporating lightweight task heads, this collaborative training strategy is designed to enhance the representational capabilities of the underlying vision backbone.}
  \vspace{-1em}
 \label{fig:arch}
\end{figure*}

This section starts by formulating our considered problem in Section~\ref{problem}; 
then we elaborate on the overall multi-task collaborative post-training framework in Section~\ref{framework}; lastly, we detail the training strategy of VersaViT in Section~\ref{training_strategy}.

\vspace{2pt}

\subsection{Problem Formulation}
\label{problem}

Our goal is to strengthen the MLLM vision backbone so that its representations support both language-mediated reasoning and dense, pixel-level perception. To this end, we propose VersaViT, a well-rounded vision backbone that is optimized by a multi-task collaborative post-training framework. This framework augments a shared vision encoder with lightweight, task-specific heads, enabling multi-granularity supervision with minimal architectural change.
As shown in Figure~\ref{fig:arch}, we select three representative tasks: (i) VQA and image captioning for semantic alignment with language, (ii) monocular depth estimation for 3D-aware spatial reasoning, and (iii) image referring segmentation for localized, language-conditioned perception.

Our framework uses separate data streams for each task. To simplify, we illustrate this using a single image for all tasks. Given an input image $\boldsymbol{I}\in\mathbb{R}^{H\times W\times 3}$, the shared vision encoder ($\mathrm{\Phi}_V$, parameterized by $\boldsymbol{\theta}_V$) first extracts the corresponding visual features~($\{\mathcal{F}_{i}^{V}\}_{i=1}^{N}$), where $N$ denotes the number of the transformer layers within the vision encoder. The extracted features are subsequently routed to task-specific heads with parameters~$\boldsymbol{\theta}_{\text{heads}}$. The resulting respective task losses ($\mathcal{L}_{\text{cap}}$, $\mathcal{L}_{\text{depth}}$, $\mathcal{L}_{\text{seg}}$) are aggregated to form a combined objective~$\mathcal{L}_{\text{all}}$, which is backpropagated to collaboratively optimize and strengthen the shared vision backbone ($\mathrm{\Phi}_V$) across all objectives. The overall optimization goal is to find the optimal parameters for both the backbone and the heads by minimizing the combined loss: 

$$\min_{\boldsymbol{\theta}_V, \boldsymbol{\theta}_{\text{heads}}} \mathcal{L}_{\text{all}} = \sum_{\tau \in \mathcal{T}} \lambda_{\tau} \mathcal{L}_{\tau}(\boldsymbol{\theta}_{V}, \boldsymbol{\theta}_{\tau}), $$
where $\mathcal{T} = \{\texttt{cap}, \texttt{depth}, \texttt{seg}\}$ is the set of all tasks, $\lambda_{\tau}$ is the task weight hyperparameter, and $\boldsymbol{\theta}_{\tau}$ represents the parameters for the head of task $\tau$.

\subsection{Overall Framework}
\label{framework}

In this section, we provide a detailed breakdown of our proposed multi-task collaborative training framework by successively introducing the three auxiliary training tasks: VQA and image captioning, monocular depth estimation, and image referring segmentation.

% \weidi{you have repeatedly described how the input image is fed into feature extractor, I would suggest to do this once at the beginning, each of the following heads simply use this feature.}

\subsubsection{Generative Vision-Language Fine-Tuning}
\label{task:vqa}

To enhance the high-level semantic understanding of the vision backbone, the combined VQA and image captioning tasks are integrated.
This section details the components and training for the VQA\&image captioning tasks.

\vspace{2pt}
\noindent \textbf{Core Components.} To accomplish the tasks of VQA and image captioning, we adopt the mainstream MLLM architecture~\cite{liu2023visual}. 
Specifically, our main components consist of a vision encoder~(\texttt{Qwen2-VL-ViT}), a vision projector~(an MLP Layer), and a large language model~(\texttt{Qwen3-1.7B}).

\vspace{2pt}
\noindent \textbf{Training Pipeline.} For the VQA and image captioning task, we utilize the final layer feature~($\mathcal{F}_{N}^{V}$) from the vision encoder. Given an input image $\boldsymbol{I}$ and an associated text~$\boldsymbol{t} = (t_0, t_1, \cdots, t_L)$ with the token length of $L$, the visual feature~($\mathcal{F}_{N}^{V}$) is first passed through a vision projector to obtain the projected image embeddings. These projected image embeddings are then concatenated with the input text embedding derived from $\boldsymbol{t}$ via a text embedding layer. The resulting sequence forms the full input for the LLMs. The training objective are as follows: $\mathcal{L}_{\text{cap}} = \frac{1}{L} \sum_{i=1}^{L} \mathcal{L}_{\text{ce}}(t_i |\boldsymbol{I}, t_0, \cdots, t_{i-1})$.

\vspace{2pt}
\noindent \textbf{Captioning Alignment Warm-Up.} Prior to the multi-task collaborative training phase, we employ the captioning task to warm up the vision projection. Crucially, during this stage, both the vision backbone and the LLM are frozen; we exclusively train the vision projection using autoregressive loss~($\mathcal{L}_{\text{cap}}$). For training data, we incorporate a variety of sources, including natural image-text pairs~({\em e.g.}, pixmo-cap~\cite{deitke2025molmo} and SA-1B-InternVL~\cite{kirillov2023segment}) and optical character recognition (OCR) data~({\em e.g.}, idl-wds~\cite{biten2022ocr}).

\subsubsection{Monocular Depth Estimation}
\label{subsec: depth estimation}

To address the feature deficiency in spatial awareness and improve geometric representation, the monocular depth estimation task is incorporated. This section introduces the methodology for obtaining the depth prediction and corresponding pseudo ground truth label for model training, as well as the training objective employed. 

\vspace{2pt}
\noindent \textbf{Depth Prediction.} For monocular depth estimation, we select a uniform subset of features from the extracted visual features ($\{\mathcal{F}_{i}^{V}\}_{i=1}^{N}$). Specifically, we uniformly sample three of these features and feed them into a DPT head~\cite{ranftl2021vision} to generate the final depth prediction~($d^*\in\mathbb{R}^{H\times W}$).

\vspace{2pt}
\noindent \textbf{Pseudo Depth Labeling.} We do not rely on manually annotated ground truth data. Instead, we employ a pseudo-labeling strategy to generate the depth labels. Specifically, we adopt the Depth Anything V2 model~\cite{yang2024depth} to annotate the images to get the depth labels~($d\in\mathbb{R}^{H\times W}$). Details regarding the data sources are provided in Table~\ref{tab:dataset}.

\vspace{2pt}
\noindent \textbf{Training Objective.} We use a scale- and shift-invariant loss $\mathcal{L}_{\text{ssi}}$ and a gradient matching loss $\mathcal{L}_{\text{gm}}$ to optimize depth estimation. Specifically, $\mathcal{L}_{\text{ssi}} = \frac{1}{HW}\sum_{i=1}^{HW}\rho(d_i^*, d_i)$, where $\rho$ is the affine-invariant mean absolute error loss. $\mathcal{L}_{\text{gm}}$ evaluate the gradient difference between the ground truth and the rescaled estimates on multiple scales $k$: 

\vspace{-8pt}
$$\mathcal{L}_{\text{gm}} = \sum_{k=1}^K  \sum_{i=1}^{HW} | s \nabla_x^k d_i - \nabla_x^k d_i^* | + | s \nabla_y^k d_i - \nabla_y^k d_i^* |.$$

The final loss function is given by: $\mathcal{L}_{\text{depth}} = \lambda_{\text{ssi}}\mathcal{L}_{\text{ssi}} + \lambda_{\text{gm}}\mathcal{L}_{\text{gm}}$, where $\lambda_{\text{ssi}}$ and $\lambda_{\text{gm}}$ are weight hyperparameters. Please refer to the supplementary materials for more details.

\subsubsection{Image Referring Segmentation}

To instill robust localization and pixel-level fine-grained perception into the visual representation, the image referring segmentation task is employed. This section outlines the way to get the mask prediction, the data scaling strategy, and the corresponding training objective.

\vspace{2pt}
\noindent \textbf{Mask Prediction.} Given an input image $\boldsymbol{I}$ and an associated text prompt~$\boldsymbol{t}$, we use the final layer feature~($\mathcal{F}_{N}^{V}$) from the vision encoder. Subsequently, we utilize a pretrained text embedding model~(\texttt{Qwen3-Embedding-8B~\cite{zhang2025qwen3}}) as the text encoder to offline extract the corresponding text embedding. Furthermore, this text embedding is then processed by the prompt encoder to yield the prompt embedding~($\mathcal{F}^{P}$). Ultimately, the mask decoder efficiently mapping $\mathcal{F}_{N}^{V}$, $\mathcal{F}^{P}$, and an output token to generate the final mask~($\hat M$). For further details about the architecture of the prompt encoder and mask decoder, please refer to~\cite{kirillov2023segment}.

\vspace{2pt}
\noindent \textbf{Segmentation Data Construction.} To efficiently scale the segmentation data,  which is structured as an image $\boldsymbol{I}$ associated with a set of $P$ instance-level pairs $\{(\boldsymbol{t}_i, M_i)\}_{i=1}^{P}$~(where $\boldsymbol{t}_i$ is the text description and $M_i$ is the corresponding mask), we employ the following three strategies: (i) We directly integrate existing referring segmentation datasets. (ii) For datasets already possessing mask annotations ({\em e.g.}, SA-1B), we leverage a dense captioning model~({\em e.g.}, Describe Anything~\cite{lian2025describe}) to generate a text annotation for each mask. (iii) For datasets annotated only with bounding boxes and class labels, we first use SAM~\cite{kirillov2023segment} to generate precise masks based on the bounding boxes, and subsequently apply the Describe Anything or the existing class label for text description annotation. For detailed data regarding segmentation, please refer to Table~\ref{tab:dataset}.

% \weidi{need to give a bit of notations, for example, your data consists of image, text description, mask, is it only one language description per image ? meaning only one mask ?}

\vspace{2pt}
\noindent \textbf{Training Objective.} We utilize a combination of per-pixel binary cross-entropy (BCE) loss and DICE loss, with corresponding loss weights $\lambda_{\text{bce}}$ and $\lambda_{\text{dice}}$. Given the ground-truth target masks~($M$), the overall loss is formulated as $\mathcal{L}_{\text{seg}} = \lambda_{\text{bce}}\mathcal{L}_{\text{bce}}(\hat M, M) + \lambda_{\text{dice}}\mathcal{L}_{\text{dice}}(\hat M, M)$.

\subsection{Model Training}
\label{training_strategy}

In this section, we describe the details of our multi-task collaborative training, which uses diverse multi-task objectives to advance the capabilities of the vision backbone.

\vspace{2pt}
\noindent \textbf{Multi-task Collaborative Training.} After captioning alignment, we adopt the multi-task collaborative training schema involving the three objectives detailed in Section~\ref{framework}: VQA\&captioning~($\mathcal{L}_{\text{cap}}$), monocular depth estimation~($\mathcal{L}_{\text{depth}}$), and image referring segmentation~($\mathcal{L}_{\text{seg}}$). During this phase of training, the weights of the vision encoder, the task-specific heads, and the LLM are all unfrozen and updated. We perform joint optimization via alternating batch sampling combined with gradient accumulation. Specifically, we alternately sample a batch of data from each of the three tasks for a forward pass. The loss of each task is used to compute and accumulate gradients, with the collective parameter update executed after iterating through all three tasks. The final loss is a weighted sum:

\vspace{-4pt}
$$\mathcal{L}_{\text{all}} = \lambda_{\text{cap}}\mathcal{L}_{\text{cap}} + \lambda_{\text{depth}}\mathcal{L}_{\text{depth}} + \lambda_{\text{seg}}\mathcal{L}_{\text{seg}},$$where $\lambda_{\text{cap}}$, $\lambda_{\text{depth}}$ and $\lambda_{\text{seg}}$ are weight hyperparameters. Jointly optimizing these objectives enables the vision backbone to capture high-level semantics and strengthen pixel-level and spatial-aware representations.

%% file: sec/5_experiments.tex
\section{Experiments}

\subsection{Experimental Setup}
\label{sec:exp_setup}

\noindent \textbf{Training Datasets.} Our training is supported by a diverse dataset spanning varied sources. Specifically, the captioning alignment stage primarily utilizes high-quality image-caption and OCR data. Subsequently, during the multi-task collaborative training phase, we either collect or synthesize new data tailored to each respective task. The comprehensive details of the training data are presented in Table~\ref{tab:dataset}.

\input{tables/training_datasets}

\input{tables/open_compass}

\input{tables/dense_features_all}

\vspace{2pt}
\noindent \textbf{Evaluation Settings.} To comprehensively assess the performance of VersaViT, we conduct evaluations across multiple downstream tasks. (i) VQA performance: To evaluate VersaViT's performance on VQA, we adopt an MLLM evaluation strategy. Specifically,  we integrate VersaViT with a vision projection layer and an LLM~(\texttt{Qwen3-8B}). The training is performed in two stages: in the first stage, only the vision projection layer is trained; in the second stage, both the vision projection layer and the LLM are trained. Throughout this entire training process, the parameters of the vision encoder are kept frozen. For details regarding the training data utilized for VQA evaluation, please refer to the supplementary materials. (ii) Dense Feature Probing: To demonstrate VersaViT's capability for dense feature representation, we follow the methodology of~\cite{simeoni2025dinov3} by employing a linear probing approach with the vision backbone kept frozen. (iii) In-Task and Out-of-Task Performance: To assess our model's performance on the tasks included in the multi-task collaborative training stage, we conduct evaluations using established benchmarks specific to those tasks and compare the results against specialized models. Furthermore, for out-of-task generalization, we evaluate VersaViT's retrieval performance on corresponding benchmarks.

\vspace{2pt}
\noindent \textbf{Implementation Details.} Our framework is built on PyTorch and is initialized by Qwen2-VL-7B-ViT~\cite{wang2024qwen2}. Our model is capable of accepting dynamic resolutions. In the captioning alignment stage, we conduct experiments on 128 H20 GPUs with a batch size of 1024 and a learning rate of 1e-3, training for one epoch. During the multi-task collaborative training phase, the learning rate for the vision encoder, the vision projection, and the text decoder is set to 1e-5, while the learning rate for the remaining parameters is set to 1e-4, again for one epoch. For more implementation details, please refer to the supplementary materials.

\subsection{Experimental Results}
\label{exp:results}

In this section, we compare VersaViT with the Qwen2-VL-ViT baseline across both VQA and classic vision-centric tasks to demonstrate the effectiveness of our method.

\vspace{2pt}
\noindent \textbf{Improved VQA Performance.} As outlined in Section~\ref{sec:exp_setup}, we adopt a two-stage training paradigm based on MLLMs to evaluate our vision backbone on various VQA tasks. Table~\ref{tab:opencompass} presents the performance of our model on the OpenCompass benchmark. The results indicate that (i) Crucially, under a fair comparison setting, our model demonstrates a significant improvement over the Qwen2-VL-ViT~\cite{wang2024qwen2} baseline, achieving an average gain of 1.6 points across eight benchmarks. This consistent gain strongly suggests that our backbone's high-level semantic representations are superior to the baseline. (ii) When VersaViT is coupled with an LLM through instruction tuning with small-scale data, it generally outperforms several well-established open-source models, such as MiniCPM-V-2.6~\cite{yao2024minicpm} and DeepSeek-VL2~\cite{wu2024deepseek}. However, a performance gap remains when compared to top-tier MLLMs, such as Qwen2.5-VL-7B~\cite{bai2025qwen2}, primarily due to differences in data scale. For further details regarding the data and training for these two stages, please refer to the supplementary materials.

\input{tables/task_ablation}

\vspace{2pt}
\noindent \textbf{Improved Dense Features.} To evaluate the dense features of our method, we utilize the frozen patch features from the final layer and assess their performance through linear probing on both semantic segmentation and monocular depth estimation tasks. The experimental results, as shown in Table~\ref{tab:results-linear-probing}, reveal that (i) VersaViT demonstrates superior performance to SigLIP~2 on both tasks, although there remains a slight gap compared to DINOv3. (ii) In particular, our method delivers substantial improvements over the Qwen2-VL-ViT baseline. For semantic segmentation, the performance increases from 33.6 to 49.6 on ADE20k and from 67.5 to 86.6 on Pascal VOC. For monocular depth estimation, the RMSE on the KITTI dataset is dramatically reduced from 3.735 to 3.136, and on the NYUv2 dataset, it drops from 0.541 to 0.473. These results demonstrate that our method significantly enhances the representational quality of dense features within the vision backbone.

% \vspace{2pt}
% \noindent \textbf{Overall Summary.}

% \input{tables/linear_probing_depth_exp}

% \begin{figure}[htbp]
%   \centering
%   \includegraphics[width=.47\textwidth]{pics/perception_vqa_qirui_latest.pdf} \\
%   \vspace{-0.2cm}
%   \caption{\textbf{Performance  on perception-focused VQA benchmarks.} $*$ denotes the average of CVBench-2D and CVBench-3D.}
%   \vspace{-1em}
%  \label{fig:perception}
% \end{figure}
\input{tables/loss_weight_ablation}
\input{supp_tables/backbone_ablation}
\input{tables/method_data_vs}
\input{tables/3D_correspondence}
\input{tables/depth_estimation}
\input{tables/image_referring_seg}

\subsection{Ablation Study \& Analysis}
\label{sec:ablation}

In this section, we further investigate the effect of our method, for example, the effectiveness of multi-task training, the generalization of our method, {\em etc}. 

\vspace{2pt}
\noindent \textbf{Effectiveness of Multi-task Training.} As shown in Table~\ref{tab:ablation_task}, it can be observed that as the number of tasks during the multi-task collaborative training phase increases, the model's overall performance improves. The following key observations can be made: (i) Training only the VQA and Image Captioning tasks effectively enhances the model's performance on the VQA benchmark, but it degrades the vision backbone's dense feature representation. This finding aligns with the observation made in Section~\ref{sec:motivation}. (ii) Jointly training the VQA and depth tasks significantly improves the model's performance on both VQA and depth-related tasks. (iii) Adding the segmentation task further enhances the model's capabilities, not only improving segmentation performance but also boosting performance in both VQA and depth tasks. Collectively, integrating these three tasks results in a more comprehensive visual representation.

\vspace{2pt}
\noindent \textbf{Ablation Study on Loss Weights.} As shown in Table~\ref{tab:ablation_loss_weight}, we conduct ablation experiments under different loss-weight settings. The experimental results indicate that increasing the loss weight of a specific task generally leads to slightly improved performance on that task. Moreover, balanced weights achieve the optimal trade-off among all tasks. Importantly, our method is robust to weight variations, as all configurations significantly outperform the baseline.

\vspace{2pt}
\noindent \textbf{Generalization on Different Vision Backbones.} In this section, we investigate the application of our multi-task post-training framework across different vision backbones. As shown in Table~\ref{tab:supp_vision_backbone}, our approach is compatible with various ViT architectures and consistently enhances the performance of the backbones across multiple aspects.

% \vspace{2pt}
% \noindent \textbf{Analysis of Perception-focused VQA Benchmarks.} The analysis presented in Section~\ref{exp:results} demonstrates that VersaViT possesses improved dense features. Following this observation, we investigate whether the enhancement of ViT's vision-centric dense features translates into performance gains on perception-focused VQA benchmarks. As illustrated in Figure~\ref{fig:perception}, an evaluation is conducted using four VQA benchmarks that prioritize perception. The results indicate that our proposed model achieves notable performance improvements across all these benchmarks.

\vspace{2pt}
\noindent \textbf{Data-driven Vs. Method-driven Gains.} To isolate the effect of multi-tasking from data scaling, we introduce a baseline model trained on a 35M-scale dataset using only captioning and VQA tasks, excluding the multi-tasking framework. As demonstrated in Table~\ref{tab:multi_tasking_vs_data}, this comparison confirms that the performance gains stem from our multi-tasking methodology rather than mere data scaling.

\vspace{2pt}
\noindent \textbf{Enhanced 3D Correspondence.} We follow the evaluation setting established by Probe3D~\cite{el2024probing} to assess geometric correspondence on the NAVI dataset~\cite{jampani2023navi} and semantic correspondence on the SPair dataset~\cite{min2019spair}. As demonstrated in Table~\ref{tab:3d_correspondence}, VersaViT significantly outperforms the baseline, achieving a 2.3-point increase on NAVI and a significant 10-point increase on SPair. This compelling evidence strongly indicates that our method effectively enhances the 3D-aware representations of the vision backbone. Further details are provided in the supplementary materials.

% \begin{figure*}[ht]
%   \centering
%   \includegraphics[width=\textwidth]{pics/visualize.pdf} \\
%   \vspace{-6pt}
%   \caption{\textbf{Qualitative examples.} Qualitative comparison of our model against Qwen2VL-ViT baseline across three tasks: VQA, semantic segmentation, and depth estimation. For the VQA task, we conduct the evaluation using Qwen3-8B. The results for semantic segmentation and depth estimation are obtained through linear probing. As can be observed, our method outperforms Qwen2-VL-ViT in all these tasks.}
%   \vspace{-1em}
%  \label{fig:visualize}
% \end{figure*}

\vspace{2pt}
\noindent \textbf{Transfer to Tasks Learned During Multi-task Training.} Through the multi-task training stage, our model naturally acquires the ability to perform monocular depth estimation and image referring segmentation. For monocular depth estimation, we evaluate our model on three benchmarks, with the results detailed in Table~\ref{tab:monocular_depth_estimation}. Despite being trained solely via multi-task learning, our method achieves performance on par with that of specialized depth estimation models. Moreover, after subsequent fine-tuning specifically for the depth estimation task, our model attains performance comparable to state-of-the-art methods, demonstrating that VersaViT is highly adaptable to the depth estimation task.

\input{tables/retrieval}

Similarly, for image referring segmentation, the experimental results are shown in Table~\ref{tab:image_ref}. After fine-tuning the multi-task pre-trained model on the RefCOCO datasets, our model substantially outperforms a wide range of specialized methods, demonstrating strong transferability.

\vspace{2pt}
\noindent \textbf{Transfer to Retrieval Tasks.} To assess the performance of our trained vision backbone on retrieval tasks, we select \texttt{siglip2-so400m-patch14-384} as the text encoder. We conduct two experimental settings: a frozen configuration, where only the vision attention pooling layer and the text projection are trained; and a more extensive setting where the vision encoder training is enabled to fully explore the potential of our model. For training data, we use the CC3M, CC12M, and YFCC15M datasets, as used in~\cite{zheng2024dreamlip}. For further details regarding the data and architecture, please refer to the supplementary materials.

The experimental results, presented in Table~\ref{exp:retrieval}, show that our model exhibits exceptional retrieval performance across various settings. Notably, with only 30M training samples and training solely the lightweight projection layer, our model surpasses CLIP and achieves comparable performance to DINOv3. Furthermore, when the vision encoder is enabled, our model performs slightly behind SigLIP~2 by about 2 points. This outcome clearly indicates that the vision encoder trained using our framework still demonstrates strong potential and competitiveness in retrieval tasks.

\begin{figure}[ht]
  \centering
  \includegraphics[width=.5\textwidth]{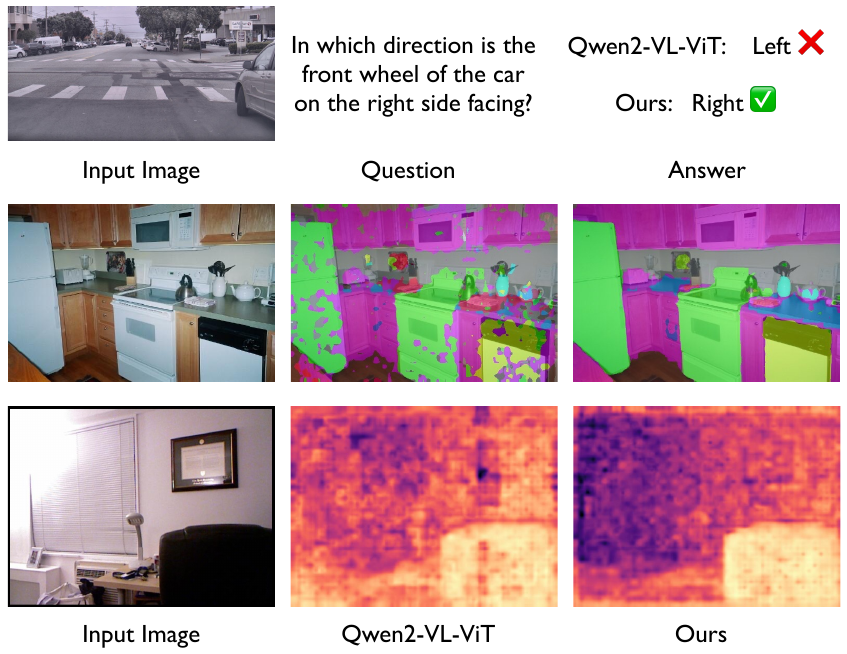} \\
  \vspace{-6pt}
  \caption{\textbf{Qualitative examples.} Qualitative comparison of VersaViT against Qwen2-VL-ViT across three tasks. For the VQA, we evaluate using Qwen3-8B. The results for semantic segmentation and depth estimation are obtained through linear probing. As shown, our method outperforms the baseline in all these tasks.}
  \vspace{-0.4cm}
 \label{fig:visualize}
\end{figure}

\vspace{2pt}
\noindent \textbf{Discussion on Seamless Integration.} While our multi-task collaborative training method is validated via post-training on the vision encoder, its design enables seamless integration into any stage of MLLM training ({\em e.g.}, pre-training, decay, and post-training). Typically, MLLM training relies exclusively on autoregressive loss; in contrast, our method augments this with two pixel-level objectives. This integration is facilitated by batch-wise task alternation, wherein only one task-specific loss is computed per batch. Such a design ensures that our method can be seamlessly adapted to existing MLLM training frameworks. Moreover, we anticipate that applying our approach during the earlier phases of MLLM training will allow the vision backbone to acquire more precise and semantically rich feature representations, which we consider an avenue for future work.

\subsection{Qualitative Results}

% In Figure~\ref{fig:visualize}, we provide a qualitative comparison of our method with the Qwen2VL-ViT baseline across three distinct tasks: VQA, semantic segmentation, and depth estimation. As demonstrated, after undergoing multi-task collaborative training, our enhanced vision backbone significantly outperforms the baseline in each of these tasks. For instance, as shown in the first row, our model is able to more accurately distinguish object size and better understand spatial relationships. The examples in the second row indicate that our model achieves more precise overall object segmentation, for instance, by accurately segmenting complete entities such as refrigerators and windows. Furthermore, the examples in the third row demonstrate that our model can more effectively recognize the depth of objects and spatial relationships, such as the spatial separation between a person and a chair. Collectively, these results highlight our model's enhanced ability to capture fine-grained features, resulting in more precise performance in VQA, object segmentation, and depth estimation.

In Figure~\ref{fig:visualize}, we provide a qualitative comparison of VersaViT with the Qwen2-VL-ViT baseline across three tasks: VQA, semantic segmentation, and depth estimation. As demonstrated, after undergoing multi-task collaborative training, VersaViT significantly outperforms the baseline in each of these tasks. For instance, the example in the first row illustrates that our model exhibits a superior comprehension of spatial relationships. Furthermore, the examples in the second and third rows highlight our model's enhanced ability to capture fine-grained features, resulting in more precise object segmentation and depth estimation.

%% file: tables/training_datasets.tex
\begin{table}[t]
\centering
% 使用 adjustbox 的 resizebox，并设置 \tabcolsep 让列之间稍微紧凑
\caption{\textbf{Statistics of the training data.} This table summarizes the datasets utilized in our model training.}
\vspace{-0.2cm}
\begin{adjustbox}{width=0.48\textwidth} 
\setlength{\tabcolsep}{4pt} % 适当增加列间距，避免内容过于拥挤
\begin{tabular}{llrr} % l:左对齐, r:右对齐 (Data Size)
\toprule
\textbf{Training Stage} & \textbf{Task} & \textbf{Data Source} & \textbf{Data Size} \\
\midrule
\multirow{3}{*}{\shortstack{Captioning \\ Alignment}} % 使用 \shortstack 简化换行和居中
  & \multirow{3}{*}{Image Captioning} & Pixmo-cap~\cite{deitke2025molmo} & 0.6M\\
  & & IDL-WDS~\cite{biten2022ocr} & 4.7M \\
  & & SA-1B-InternVL~\cite{kirillov2023segment} & 4.1M \\
\midrule % 粗一些的midrule，用于 Training Stage 的大分组

\multirow{10}{*}{\shortstack{Multi-task \\ Collaborative \\ Training}} 
  & \multirow{6}{*}{Depth Estimation} & Open-images~\cite{kuznetsova2020open} & 1.7M\\
  & & Objects365~\cite{shao2019objects365} & 1.7M \\
  & & Google-landmark~\cite{weyand2020google} & 2.7M \\
  & & Places365~\cite{zhou2017places} & 8.0M \\ % 统一小数点位数
  & & SA-1B~\cite{kirillov2023segment} & 11.0M \\ % 统一小数点位数
  & & CC3M~\cite{changpinyo2021conceptual} & 2.8M \\
  
  \cmidrule(lr){2-4} % 较细的分隔线，仅分隔 Task 组
  
  & \multirow{3}{*}{Referring Segmentation} & GRIT~\cite{peng2023kosmos} & 13.9M \\
  & & Objects365~\cite{shao2019objects365} & 1.6M \\
  & & SA-1B-DAM~\cite{lian2025describe} & 0.6M \\
  
  \cmidrule(lr){2-4} % 较细的分隔线
  
  & VQA\&Image Captioning & FineVision~\cite{wiedmann2025finevision} & 14.0M \\
\bottomrule
\end{tabular}
\end{adjustbox}
\vspace{-1em}
\label{tab:dataset}
\end{table}

%% file: tables/open_compass.tex
\begin{table*}[t]
\centering 
\caption{\textbf{Comparison between different methods on OpenCompass benchmarks, a collection of eight general VQA benchmarks.}We assess our method on the following benchmarks, MMBench~\cite{liu2024mmbench}, MMstar~\cite{chen2024we}, MMMU~\cite{yue2024mmmu}, MathVista~\cite{lu2023mathvista}, HallusionBench~\cite{guan2024hallusionbench}, AI2D~\cite{kembhavi2016diagram}, OCRBench~\cite{liu2023hidden}, MMVet~\cite{yu2023mm}. For details on our VQA training and data, please refer to the supplementary materials.}
\vspace{-0.2cm}
\footnotesize 
\setlength{\tabcolsep}{6pt} 
\renewcommand{\arraystretch}{1.2} 
\begin{tabular}{l c c c c c c c c c}
\toprule
\textbf{Method} & \textbf{MMBench} & \textbf{MMStar} & \textbf{MMMU} & \textbf{MathVista} & \textbf{HallusionBench} & \textbf{AI2D} & \textbf{OCRBench} & \textbf{MMVet} & \textbf{Avg.} \\
% \midrule
% \multicolumn{10}{l}{\textbf{Proprietary Models}} \\
% GPT-5~\cite{openai_gpt5_systemcard} & 86.6 & 75.7 & 81.8 & 81.9 & 65.2 & 89.5 & 80.7 & 77.6 & 79.9 \\ 
% Gemini-2.5-Pro~\cite{comanici2025gemini} & 88.3 & 73.6 & 74.7 & 80.9 & 64.1 & 89.5 & 86.2 & 83.3 & 80.1 \\ 

\midrule
\multicolumn{10}{l}{\textbf{Open-Source Models (without access to the training data)}} \\ 
% LLaVA-OneVision-7B~\cite{li2024llava} & 76.8 & 56.7 & 46.8 & 58.6 & 47.5 & 82.8 & 69.7 & 50.6 & 61.2 \\ 
MiniCPM-V-2.6~\cite{yao2024minicpm} & 78.0 & 57.5 & 49.8 & 60.8 & 48.1 & 82.1 & 85.2 & 60.0 & 65.2 \\
DeepSeek-VL2~\cite{wu2024deepseek} & 81.2 & 61.9 & 54.0 & 63.9 & 45.3 & 83.8 & 80.9 & 60.0 & 66.4 \\
Qwen2-VL-7B~\cite{wang2024qwen2} & 81.0 & 60.7 & 53.7 & 61.6 & 50.4 & 83.0 & 84.3 & 61.8 & 67.1 \\
Qwen2.5-VL-7B~\cite{bai2025qwen2} & 82.2 & 64.1 & 58.0 & 68.1 & 51.9 & 84.3 & 88.8 & 69.7 & 70.9 \\
% InternVL3-8B~\cite{zhu2025internvl3} & 82.1 & 68.7 & 62.2 & 70.5 & 49.0 & 85.1 & 88.4 & 82.8 & 73.6 \\
\midrule
\multicolumn{10}{l}{\textbf{Ours (LLM = Qwen3-8B)}} \\ 
\rowcolor{green!10} Qwen2-VL-ViT~\cite{wang2024qwen2} & 78.2 & 59.3 & 51.0 & 62.1 & 49.0 & 79.4 & 80.6 & 58.7 & 64.8 \\ 
\rowcolor{green!10} & 78.0 & 60.9 & 53.1 & 63.7 & 51.3 & 79.1 & 82.2 & 62.6 & 66.4 \\ 
\rowcolor{green!10} \multirow{-1.8}{*}{VersaViT} & \textcolor{red}{(-0.2)} & \textcolor{darkgreen}{(+1.6)} & \textcolor{darkgreen}{(+2.1)} & \textcolor{darkgreen}{(+1.6)} & \textcolor{darkgreen}{(+2.3)} & \textcolor{red}{(-0.3)} & \textcolor{darkgreen}{(+1.6)} & \textcolor{darkgreen}{(+3.9)} & \textcolor{darkgreen}{(+1.6)} \\ 
\bottomrule
\end{tabular}
\label{tab:opencompass}
\end{table*}

%% file: tables/dense_features_all.tex
\begin{table}[t]
    \centering
    \caption{
        \textbf{Linear probing results on semantic segmentation and monocular depth estimation with frozen backbones.} We report the mean Intersection-over-Union (mIoU) metric for ADE20k, Cityscapes, and VOC. We report the Root Mean Squared Error (RMSE) metric for the depth benchmarks NYUv2 and KITTI. For segmentation, we use an image resolution of $560 \times 560$.
    }
    \vspace{-0.2cm}
    \footnotesize
    \setlength{\tabcolsep}{1pt}
    \begin{tabular}{l c c c c c c c}
        \toprule
        & & \multicolumn{3}{c}{\textbf{Segmentation}} && \multicolumn{2}{c}{\textbf{Depth}}\\ 
        \cmidrule{3-5} \cmidrule{7-8}
        \textbf{Method} & \textbf{Arch.} & \textbf{ADE20k} & \textbf{Citysc.} & \textbf{VOC} && \textbf{NYUv2} $\downarrow$ & \textbf{KITTI} $\downarrow$  \\
        \midrule
        \multicolumn{4}{l}{\textbf{Self-supervised Backbone}} \\
        MAE~\cite{he2022masked} & H/14 & 33.3 & 58.4 & 67.6 & & 0.517 & 3.660 \\
        % DINOv2~\cite{oquab2023dinov2} & G/14 & 49.5 & 75.6 & 83.1 & & 0.372 & 2.624\\
        Web-DINO~\cite{fan2025scaling} & 7B/14 & 42.7 & 68.3 & 76.1 & & 0.466 & 3.158 \\
        DINOv3~\cite{simeoni2025dinov3} & 7B/16 & 55.9 & 81.1 & 86.6 & & 0.309 & 2.346 \\
        \midrule % 使用 \midrule 明确分隔两大组
         \multicolumn{4}{l}{\textbf{Weakly-supervised Backbone}} \\
        OpenCLIP~\cite{cherti2023reproducible} & G/14 & 39.3 & 60.3 & 71.4 & & 0.541 & 3.570 \\
        SigLIP~2~\cite{tschannen2025siglip} & G/16 & 42.7 & 64.8 & 72.7 & & 0.494 & 3.273 \\
        PEcore~\cite{bolya2025perception} & G/14 & 38.9 & 61.1 & 69.2 & & 0.590 & 4.119 \\
        \midrule
        \multicolumn{4}{l}{\textbf{Ours}} \\
        % 保持行着色以突出“我们”的模型
        \rowcolor{green!10} Qwen2-VL-ViT~\cite{wang2024qwen2} & H/14 & 33.6 & 57.6 & 67.5 && 0.541 & 3.735 \\
        \rowcolor{green!10} & & 49.6 & 74.5 & 86.6 && 0.473 & 3.136\\
        \rowcolor{green!10} \multirow{-1.8}{*}{VersaViT} & \multirow{-1.8}{*}{H/14} & \textcolor{darkgreen}{(+16.0)} & \textcolor{darkgreen}{(+16.9)} & \textcolor{darkgreen}{(+19.1)} && \textcolor{darkgreen}{(-0.068)} & \textcolor{darkgreen}{(-0.599)} \\
        \bottomrule
    \end{tabular}
    % }
    \vspace{-0.5cm}
    \label{tab:results-linear-probing}
\end{table}

%% file: tables/task_ablation.tex
\begin{table}[t]
\centering
\caption{\textbf{Ablation study on multi-task training.} Here, we report the average score on the OpenCompass, the RMSE on the NYUv2, and the MIoU on the ADE20k.}
\vspace{-0.2cm}
\footnotesize
% \resizebox{.45\textwidth}{!}{ 
\setlength{\tabcolsep}{1.2mm}{
  \begin{tabular}{cccccc}
    \toprule
    \textbf{VQA\&Cap.} & \textbf{Depth} & \textbf{Seg} & \textbf{OpenCompass}$\uparrow$ & \textbf{NYUv2}$\downarrow$ & \textbf{ADE20k}$\uparrow$ \\
    \midrule
     \ding{56} & \ding{56} & \ding{56} & 64.8 & 0.541 & 33.6 \\
     \ding{52} & \ding{56} & \ding{56} & 66.1 & 0.557 & 32.0 \\
     \ding{52} & \ding{52} & \ding{56} & 66.3 & 0.495 & 35.1 \\
     \ding{52} & \ding{52} & \ding{52} & 66.4 & 0.473 & 49.6 \\
    \bottomrule
  \end{tabular}
}
% }
\vspace{-0.5cm}
\label{tab:ablation_task}
% \vspace{-2em}
\end{table}

%% file: tables/loss_weight_ablation.tex
\begin{table}[h]
\centering
\caption{
    \textbf{Ablation study on loss weights.}
}
\vspace{-0.2cm} 
\footnotesize
% \resizebox{.45\textwidth}{!}{ 
\setlength{\tabcolsep}{1.4mm}{
  \begin{tabular}{cccccc}
    \toprule
    \textbf{$\lambda_{\text{cap}}$} & \textbf{$\lambda_{\text{depth}}$} & \textbf{$\lambda_{\text{seg}}$} & \textbf{OpenCompass}$\uparrow$ & \textbf{NYUv2}$\downarrow$ & \textbf{ADE20k}$\uparrow$ \\
    \midrule
    \multicolumn{3}{c}{Baseline (Qwen2-VL-ViT)} & 64.8 & 0.541 & 33.6 \\ 
    \midrule
     0.50 & 0.25 & 0.25 & 66.5 & 0.470 & 48.3 \\
     0.25 & 0.50 & 0.25 & 65.7 & 0.455 & 48.9 \\
     0.25 & 0.25 & 0.50 & 65.7 & 0.476 & 50.0 \\
     0.33 & 0.33 & 0.33 & 66.4 & 0.473 & 49.6 \\
    \bottomrule
    % \vspace{-0.3cm}
  \end{tabular}
}
% }
\label{tab:ablation_loss_weight}
\vspace{-0.3cm}
\end{table}

%% file: supp_tables/backbone_ablation.tex
\begin{table}[t]
\centering
\caption{\textbf{Ablation study on different vision backbones.}}
\vspace{-0.2cm}
\footnotesize
% \resizebox{.45\textwidth}{!}{ 
\setlength{\tabcolsep}{1.3mm}{
  \begin{tabular}{llll}
    \toprule
    \textbf{Model} & \textbf{OpenCompass}$\uparrow$ & \textbf{NYUv2}$\downarrow$ & \textbf{ADE20k}$\uparrow$ \\
    \midrule
     Qwen2-VL-ViT & 64.8 & 0.541 & 33.6 \\
     VersaViT & 66.4 \textcolor{darkgreen}{(+1.6)} & 0.473 \textcolor{darkgreen}{(-0.068)} & 49.6 \textcolor{darkgreen}{(+16.0)} \\
     \midrule
     Qwen2.5-VL-ViT & 64.0 & 0.533 & 37.2 \\
     VersaViT-Qwen-2.5 & 65.2 \textcolor{darkgreen}{(+1.2)} & 0.464 \textcolor{darkgreen}{(-0.069)} & 47.2 \textcolor{darkgreen}{(+10.0)} \\
     \midrule
     Siglip2-so400M-512 & 59.3 & 0.497 & 40.9 \\
     VersaViT-Siglip2 & 60.4 \textcolor{darkgreen}{(+1.1)} & 0.481 \textcolor{darkgreen}{(-0.016)} & 49.2 \textcolor{darkgreen}{(+8.3)} \\
    \bottomrule
  \end{tabular}
}
\vspace{-0.2cm}
% }
\label{tab:supp_vision_backbone}
\end{table}

%% file: tables/method_data_vs.tex
\begin{table}[t]
\centering
\caption{\textbf{Effectiveness of multi-tasking vs. data scaling.}}
\vspace{-0.2cm}
\footnotesize
% \resizebox{.45\textwidth}{!}{ 
\setlength{\tabcolsep}{3.0mm}{
  \begin{tabular}{lccc}
    \toprule
    \textbf{Model} & \textbf{OpenCompass}$\uparrow$ & \textbf{NYUv2}$\downarrow$ & \textbf{ADE20k}$\uparrow$ \\
    \midrule
     35M-Baseline & 66.2 & 0.547 & 34.7 \\
     VersaViT & 66.4 & 0.473 & 49.6 \\
    \bottomrule
  \end{tabular}
  \vspace{-0.2cm}
}
% }
\label{tab:multi_tasking_vs_data}
\end{table}

%% file: tables/3D_correspondence.tex
\begin{table}[t]
\footnotesize
\centering
\caption{
    \textbf{Evaluation of 3D consistency of dense representations.}
}
\vspace{-0.2cm} 
\setlength\tabcolsep{3.25mm}
\begin{tabular}{@{}l  llll@{}}
    \toprule
     \multirow{2}{*}{\textbf{Method}} && \multicolumn{1}{l}{Geometric} && \multicolumn{1}{l}{Semantic} \\ 
    \cmidrule{3-5}
    && NAVI (Recall) && SPair (Recall) \\
    \midrule
    Qwen2-VL-ViT~\cite{wang2024qwen2}  && 39.27 && 17.05 \\
    VersaViT && 41.64 \textcolor{darkgreen}{(+2.37)} && 26.99 \textcolor{darkgreen}{(+9.94)} \\ 
\bottomrule
\end{tabular}
\label{tab:3d_correspondence}
\vspace{-0.2cm} 
\end{table}

% We estimate 3D keypoint correspondences across views following the evaluation protocol of Probe3D~\citep{el2024probing}.
%         To measure performance, we report the correspondence recall, \ie the percentage of correspondences falling into a specified distance.

%% file: tables/depth_estimation.tex
\begin{table}[ht]
\footnotesize
\centering
\caption{\textbf{Comparison of zero-shot relative depth estimation performance.} Ours is trained via multi-task learning without task-specific fine-tuning. $*$ denotes fine-tuning on the depth task.}
\vspace{-0.2cm}
\setlength\tabcolsep{0.9mm}
\begin{tabular}{lccccc}
\toprule
\multirow{2}{*}{\textbf{Method}} & \multicolumn{2}{c}{\textbf{KITTI}~\cite{geiger2013vision}} & \multicolumn{2}{c}{\textbf{NYUv2}~\cite{silberman2012indoor}} & \multicolumn{1}{c}{\textbf{DA-2K}~\cite{yang2024depth}} \\

\cmidrule(lr){2-3}\cmidrule(lr){4-5}\cmidrule(lr){6-6}

~ & AbsRel$\downarrow$ & $\delta_1$$\uparrow$ & AbsRel$\downarrow$ & $\delta_1$$\uparrow$ & Acc(\%) \\

\midrule
\multicolumn{6}{l}{\textbf{Specialist Models}} \\
DPT~\cite{ranftl2021vision} & 10.0 & 90.1 & 9.8 & 90.3 & - \\ 
MiDaS V3.1~\cite{birkl2023midas} & 12.7 & 85.0 & 4.8 & 98.0 & - \\
DepthFM~\cite{gui2025depthfm} & 8.9 & 91.3 & 5.5 & 96.3 & 85.8 \\
Marigold~\cite{ke2023repurposing} & 9.9 & 91.6 & 5.5 & 96.4 & 86.8 \\
Depth Anything V2~\cite{yang2024depth} & 7.5 & 94.8 & 4.4 & 97.9 & 97.4 \\

\midrule
Ours & 9.4 & 90.5 & 7.7 & 94.4 & 91.3 \\
Ours$^*$ & 7.1 & 94.5 & 5.3 & 96.9 & 94.2 \\
\bottomrule
\end{tabular}
\label{tab:monocular_depth_estimation}
\vspace{-0.2cm}
\end{table}

%% file: tables/image_referring_seg.tex
\begin{table}[t]
\setlength{\tabcolsep}{1.95pt}
\centering
\caption{\textbf{Comparison of image referring segmentation performance.} Results are shown for the fine-tuning setting (post-training on RefCOCO datasets).}
\vspace{-0.2cm}
\footnotesize
% \resizebox{0.48\textwidth}{!}{
\begin{tabular}{l ccc ccc cc}
\toprule
\multirow{2}{*}{\textbf{Method}} & \multicolumn{3}{c}{\textbf{refCOCO}~\cite{kazemzadeh2014referitgame}} & \multicolumn{3}{c}{\textbf{refCOCO+}~\cite{kazemzadeh2014referitgame}} & \multicolumn{2}{c}{\textbf{refCOCOg}~\cite{mao2016generation}} \\ 
\cmidrule(lr){2-4} \cmidrule(lr){5-7} \cmidrule(lr){8-9} 
 & val & testA & testB & val & testA & testB & val(U) & test(U) \\ 
 \midrule
 \multicolumn{9}{l}{\textbf{Specialist Models}} \\
 LAVT~\cite{yang2022lavt} & 72.7 & 75.8 & 68.8 & 62.1 & 68.4 & 55.1 & 61.2 & 62.1 \\
 ReLA~\cite{liu2023gres} & 73.8 & 76.5 & 70.2 & 66.0 & 71.0 & 57.7 & 65.0 & 66.0 \\
LISA-7B~\cite{lai2024lisa} & 74.1 & 76.5 & 71.1 & 62.4 & 67.4 & 56.5 & 66.4 & 68.5 \\
VISA-7B~\cite{yan2024visa} & 72.4 & 75.5 & 68.1 & 59.8 & 64.8 & 53.1 & 65.5 & 66.4 \\
Sa2VA-26B~\cite{yuan2025sa2va} & 82.5 & - & - & 78.8 & - & - & 79.7 & -\\ 
\midrule
% \textcolor{gray}{Ours (zero-shot)} & \textcolor{gray}{34.8} & \textcolor{gray}{40.9} & \textcolor{gray}{31.3} & \textcolor{gray}{33.9} & \textcolor{gray}{37.7} & \textcolor{gray}{31.0} & \textcolor{gray}{38.8} & \textcolor{gray}{39.2} \\
Ours & 78.8 & 81.2 & 76.2 & 69.4 & 75.0 & 63.9 & 72.0 & 74.3 \\ 
\bottomrule
\end{tabular}
% }
% \vspace{-2mm}
\label{tab:image_ref}
\vspace{-1em}
\end{table}

%% file: tables/retrieval.tex
\begin{table}[htbp]
\footnotesize
\centering
\caption{\textbf{Comparison of zero-shot image-text retrieval performance.} Results are evaluated under two settings: (i) fine-tuning only the projection layers, and (ii) unfreezing the vision backbone to fully exploit the retrieval potential. $*$ denotes the second setting.}
\vspace{-0.2cm}
% \resizebox{.5\textwidth}{!}{ % I commented out the resizebox as it was incomplete and might cause issues
\setlength\tabcolsep{3mm}
\begin{tabular}{lcccc}
\toprule
\multirow{2}{*}{\textbf{Method}} & \multicolumn{2}{c}{\textbf{COCO}} & \multicolumn{2}{c}{\textbf{Flickr}} \\
\cmidrule(r){2-3} \cmidrule(l){4-5}
& T $\rightarrow$ I & I $\rightarrow$ T & T $\rightarrow$ I & I $\rightarrow$ T \\
\midrule
% \midrule was missing after the header row
% Placeholder content added:
CLIP-L~\cite{radford2021learning} & 36.5 & 56.3 & 65.2 & 85.2 \\
EVA-CLIP-8B~\cite{sun2024eva} & 53.0 & 70.3 & 80.8 & 95.6\\
SigLIP~\cite{zhai2023sigmoid} & 52.0 & 70.2 & 80.5 & 93.5 \\
SigLIP~2~\cite{tschannen2025siglip} & 55.8 & 71.7 & 85.7 & 94.9 \\
DINOv3~\cite{simeoni2025dinov3} & 45.6 & 63.7 & - & - \\
\midrule 
\multicolumn{5}{l}{\textbf{Aligned on 30M image-text pairs}} \\
Ours & 46.9 & 60.5 & 77.0 & 88.3 \\
Ours$^*$ & 54.8 & 69.5 & 82.2 & 92.5 \\
\bottomrule
\end{tabular}
\vspace{-0.3cm}
\label{exp:retrieval}
\end{table}

%% file: sec/6_conclusion.tex
\section{Conclusion}

In this paper, we have identified that the vision encoders in current MLLMs exhibit deficiencies in their dense feature representations, as manifested in their suboptimal performance on dense prediction tasks.  To address this issue, we proposed VersaViT, a well-rounded vision backbone that employs a multi-task collaborative post-training paradigm. Specifically, our approach simultaneously leverages three distinct tasks: VQA and image captioning, monocular depth estimation, and image referring segmentation. The collaborative training aims to adapt MLLM vision encoders to achieve superior performance on both VQA and vision-centric tasks, ultimately enabling a versatile VFM. Furthermore, extensive experiments on a wide range of downstream tasks have demonstrated the significant improvement achieved by our method. We anticipate that our framework will provide valuable insight into visual foundation models and stimulate further research in this area.

%% file: sec/X_suppl.tex
% \onecolumn

% {
%     \centering
%     \Large
%     \textbf{WeViT: Enhancing Vision Backbones via Task-Guided Optimization}\\
%     \vspace{0.5em}Supplementary Material \\
%     \vspace{1.0em}
% }
% % \setcounter{page}{1}
% \appendix

\clearpage
\maketitlesupplementary

% \input{tables/dense_features_all_show}

% \section{More Implementation Details}
\section{Details about the Loss of Depth Estimation}

As outlined in Section~\ref{subsec: depth estimation} of the main paper, depth estimation is optimized by employing two distinct loss functions: a scale- and shift-invariant loss ($\mathcal{L}_{\text{ssi}}$) and a gradient matching loss ($\mathcal{L}_{\text{gm}}$). Specifically, the scale- and shift-invariant loss is defined as:

$$
    \mathcal{L}_{\text{ssi}} = \frac{1}{HW}\sum_{i=1}^{HW}\rho(d_i^*, d_i),
$$
where $\rho$ is the affine-invariant mean absolute error loss: $\rho(d^*_i, d_i) = |\hat{d}^*_i - \hat{d}_i|$. Here, $\hat{d}^*_i$ and $\hat{d}_i$ represent the scaled and shifted versions of the predicted depth ($d^*_i$) and the ground truth ($d_i$), respectively. The scaling and shifting are performed by the transformation:

$$
\hat{d}_i = \frac{d_i - t(d)}{s(d)},
$$
where $t(d)$ and $s(d)$ are utilized to align the prediction and the ground truth to zero translation and unit scale. These aligning factors are defined as:

$$
t(d) = \textrm{median}(d) \quad \text{and} \quad s(d) = \frac{1}{HW}\sum_{i=1}^{HW}|d_i - t(d)|.
$$

In addition to $\mathcal{L}_{\text{ssi}}$, the gradient matching loss ($\mathcal{L}_{\text{gm}}$) evaluates the difference in the depth gradients between the ground truth and the rescaled estimates across multiple scales, indexed by $k$. This loss is formally expressed as:

$$
    \mathcal{L}_{\text{gm}} = \sum_{k=1}^K \sum_{i=1}^{HW} \left( | s \nabla_x^k d_i - \nabla_x^k d_i^* | + | s \nabla_y^k d_i - \nabla_y^k d_i^* | \right).
$$

Consequently, the final loss function for depth estimation is formulated as a weighted combination of these two terms:

$$
    \mathcal{L}_{\text{depth}} = \lambda_{\text{ssi}}\mathcal{L}_{\text{ssi}} + \lambda_{\text{gm}}\mathcal{L}_{\text{gm}},
$$
where $\lambda_{\text{ssi}}$ and $\lambda_{\text{gm}}$ denote the weight hyperparameters that govern the contribution of each loss component.

\section{More Training Details}

\subsection{More Implementation Details}

By default, we set the loss weights as follows: $\lambda_{\text{ssi}} = 1.0$, $\lambda_{\text{gm}} = 0.5$, $\lambda_{\text{bce}} = 2.0$, and $\lambda_{\text{dice}} = 0.5$. Furthermore, the weights for the various tasks are uniformly set such that $\lambda_{\text{cap}} = \lambda_{\text{depth}} = \lambda_{\text{seg}} = 1/3$. We employ distinct batch sizes for the different tasks. Specifically, the batch size for the VQA and image captioning tasks is set to 8 per card, 
while the batch size for monocular depth estimation and image referring segmentation is set to 32. In terms of input resolution, during the training of VQA and captioning, it supports a maximum image input resolution of $896 \times 896$. For the depth estimation and referring segmentation tasks, however, we resize the images to a fixed resolution for training. Images for the image referring segmentation task are resized to $560 \times 560$ pixels. In contrast, for monocular depth estimation, we select a resolution of $532 \times 532$ pixels, which is the closest size to the $518 \times 518$ used by Depth Anything V2~\cite{yang2024depth} while still being divisible by 28. 
Regarding the training data, since the dataset size varies across different tasks, we perform a resampling operation on the smaller datasets before training. 
This procedure ensures that all tasks are trained for the same number of optimization steps. We employ the Deepspeed Zero2~\cite{rajbhandari2020zero} and gradient checkpointing strategies, utilizing AdamW~\cite{loshchilov2017decoupled} optimization and selecting bf16 data precision. For linear probing in semantic segmentation, the input images are resized to $560 \times 560$ pixels. For linear probing in depth estimation, we adhere to the standard resolution specified by the corresponding dataset.

\input{supp_tables/VQA_dataset}

\subsection{More Training Data Details}

This section details the dataset used for VersaViT training. For the captioning alignment stage, we use the full Pixmo-cap dataset~\cite{deitke2025molmo}, while the IDL-WDS~\cite{biten2022ocr} and SA-1B-InternVL~\cite{kirillov2023segment} datasets are randomly sampled. For the multi-task collaborative training stage, depth-related data are annotated using \texttt{Depth-Anything-V2-Large}~\cite{yang2024depth}. For referring segmentation data, such as GRIT~\cite{peng2023kosmos}, we generate the corresponding masks using SAM~\cite{kirillov2023segment} based on the provided box annotations. For Object365~\cite{shao2019objects365}, the mask annotations are also produced by SAM from the bounding boxes, and the text annotations are obtained using \texttt{DAM-3B-Self-Contained}~\cite{lian2025describe}. The SA-1B-DAM~\cite{lian2025describe} dataset is directly usable without additional processing. For VQA and captioning data, we remove pure-text and multi-image samples from FineVision~\cite{wiedmann2025finevision} dataset and retain only the single-image instances.

\section{More Evaluation Details}

\subsection{Details about the Evaluation of VQA}

The training data for the VQA evaluation is detailed in Table~\ref{tab:supp_vqa_dataset}. In the first stage, we use a 1.7M dataset of image captions and OCR data to warm up the vision projection, with the learning rate set to 5e-4. In the second stage, we incorporate a diverse range of data types, including OCR, math, knowledge, and openQA, with a total dataset size of 6.6M. For this stage, the learning rate is set to 2e-5. For benchmark evaluation, we utilize the VLMEvalKit~\cite{duan2024vlmevalkit}.

\subsection{Details about the Evaluation of 3D Correspondence}

We adopt the evaluation setting established by DINOv3~\cite{simeoni2025dinov3}. Specifically, images are resized to a side length of 448 pixels for the NAVI dataset and 896 pixels for the SPair dataset. To measure performance, we report the correspondence recall, {\em i.e.} the percentage of correspondences falling within a specified distance. All reported results utilize features extracted from the final layer of the backbone. 

\subsection{Details about the Evaluation of Depth}

As outlined in Section~\ref{sec:ablation} of the main paper, the fine-tuning setting of depth estimation employs the same training data used in the multi-task training phase. Both the vision backbone and the DPT head are initialized with the checkpoint obtained from multi-task training. The vision backbone is optimized with a learning rate of 1e-5, whereas the remaining parameters are updated using a learning rate of 1e-4. A batch size of 32 is used per GPU, and the model is trained for 5 epochs on 128 H20 GPUs.

\subsection{Details about the Evaluation of Referring Seg}

The fine-tuning stage involves utilizing the RefCOCO series and COCO-Stuff datasets. The vision backbone and the segmentation head are both initialized using the multi-task trained checkpoint. During this phase, the parameters for the vision backbone and the segmentation head are unfrozen. For the text encoder, we employ LoRA for fine-tuning, setting both the LoRA alpha and rank to 8. The vision backbone is trained with a learning rate of 1e-5, while the remaining parameters utilize a learning rate of 1e-4. A batch size of 32 is used per GPU, and the model is trained for 50 epochs on 8 H20 GPUs.

\subsection{Details about the Evaluation of Retrieval}

Given that the Qwen2-VL-ViT lacks a CLS token, we follow the approach used in SigLIP and add an attention pooling layer at the end of the vision encoder to extract global features. This layer is specifically designed to derive a global image feature representation and comprises a multi-head attention layer followed by an MLP. Concurrently, for the text modality, we introduce a text projection layer, which is also composed of an MLP, to map the text embeddings to the same dimensionality as the image features. We specify the final shared embedding dimension to 1280. 

Regarding the training data, we follow the approach outlined in~\cite{zheng2024dreamlip}, utilizing the CC3M, CC12M, and YFCC15M datasets. Specifically, the training process incorporates the raw caption, shortsv caption, and longsv caption fields from these sources. Here, shortsv and longsv refer to short and long captions, respectively, synthesized using ShareGPT4V~\cite{chen2024sharegpt4v}.
To accelerate training, the text embeddings are extracted and prepared offline before the training phase. The retrieval experiments are conducted across 128 H20 GPUs for 10 epochs, with the batch size per GPU set to 128. When only the projection layer is fine-tuned, the learning rate is set to 5e-4. When the vision backbone is made trainable, its learning rate is set to  1e-5, while the learning rate for all other parameters is set to 1e-4.

\section{Summarization of Lightweight Task Head}

Table~\ref{tab:supp_task_head} presents the specific parameter counts of the lightweight heads used for different tasks. As shown, the parameter sizes of the task heads are minimal compared with those of the vision backbone. This design not only allows more task-related capabilities to be integrated into the vision backbone but also helps avoid excessive computational overhead.

\input{supp_tables/task_heads}
\input{supp_tables/seg_resolution}

\section{Additional Experimental Results}

% \subsection{Ablation Study on Different Vision Backbone}

% In this section, we investigate the application of our multi-task post-training framework across different vision backbones. As shown in Table~\ref{tab:supp_vision_backbone}, our approach is compatible with various ViT architectures and consistently enhances the performance of the backbones across multiple aspects.

\subsection{Ablation Study on Resolution}

As shown in Table~\ref{tab:supp_seg_resolution}, we perform linear probing experiments for semantic segmentation at multiple resolutions. The results demonstrate that our model achieves consistently and significantly higher performance than the baseline across different resolutions. Furthermore, the performance improvement increases with resolution, which may be attributed to the fact that the referring segmentation task is trained at a resolution of $560 \times 560$.

\subsection{Transfer to Classification Task}
To investigate the performance of our ViT when transferred to classification tasks, we fine-tune the model on 30M samples using the same experimental setup as in the retrieval evaluation. As shown in Table~\ref{exp:classification}, despite being trained on only 30M samples, our model exhibits strong performance on ImageNet. Moreover, on the more fine-grained MMVP benchmark, it also achieves competitive results. These findings indicate that our ViT retains strong discriminative capability for classification.

\input{supp_tables/classification}

\section{Limitation and Future Work}

Considering training efficiency, we currently set the upper bound of the training resolution to $896 \times 896$, while keeping the resolutions for the depth and segmentation tasks fixed. This configuration may cause the model to learn task-specific biases at different resolutions. In future work, we will explore applying dynamic resolutions to the depth and segmentation tasks as well. One feasible strategy is to maintain a fixed resolution with each batch while varying the resolution across batches. In addition, we plan to further expand the range of dynamic resolutions used during training to enhance the model's generalization capability.

%% file: supp_tables/VQA_dataset.tex
\begin{table}[t]
\centering
\footnotesize
% 使用 adjustbox 的 resizebox，并设置 \tabcolsep 让列之间稍微紧凑
% \begin{adjustbox}{\textwidth} 
\caption{\textbf{Data statistics for VQA evaluation.} This table summarizes the datasets utilized in the VQA evaluation stage.}
\vspace{-0.2cm}
\setlength{\tabcolsep}{2pt} % 适当增加列间距，避免内容过于拥挤
\begin{tabular}{llll} % l:左对齐, r:右对齐 (Data Size)
\toprule
\textbf{Training Stage} & \textbf{Data Type} & \textbf{Data Source} & \textbf{Data Size} \\
\midrule
\multirow{2}{*}{Pre-training} % 使用 \shortstack 简化换行和居中
  & Captioning & Laion-5B~\cite{schuhmann2022laion}, OpenVid~\cite{nan2024openvid} & 1.7M \\
  & OCR & Laion-5B~\cite{schuhmann2022laion} & 0.1M \\
\midrule % 粗一些的midrule，用于 Training Stage 的大分组

Instruction-tuning & Mixed Data & Mixed Data Source & 6.6M \\
\bottomrule
\end{tabular}
% \end{adjustbox}
\label{tab:supp_vqa_dataset}
\end{table}

%% file: supp_tables/task_heads.tex
\begin{table}[t]
\centering
\footnotesize
\caption{\textbf{Parameter statistics for various task heads.}}
\vspace{-0.2cm}
% \begin{adjustbox}{width=0.48\textwidth} 
\setlength{\tabcolsep}{2pt} % 适当增加列间距，避免内容过于拥挤
\begin{tabular}{llr} % l:左对齐, r:右对齐 (Data Size)
\toprule
\textbf{Tasks} & \textbf{Components}  & \textbf{Params.} \\
\midrule
\textcolor{gray}{-} & \textcolor{gray}{Vision Backbone} & \textcolor{gray}{644.5 M} \\
VQA\&Captioning & MLP Layer & 35.0 M \\
Referring Segmentation & Neck \& Prompt Enc. \& Mask Dec.  & 4.8 M \\
Depth Estimation & DPT & 37.0 M \\
\bottomrule
\end{tabular}
% \end{adjustbox}
% \vspace{-1em}
\label{tab:supp_task_head}
\end{table}

%% file: supp_tables/seg_resolution.tex
\begin{table}[t]
    \centering
    \footnotesize
    \caption{
        \textbf{Linear probing results on semantic segmentation with frozen backbones under different resolutions.}
    }
    \setlength{\tabcolsep}{3pt}
    % \resizebox{.48\textwidth}{!}{
    \begin{tabular}{l l lll}
        \toprule
        \textbf{Method} & \textbf{Res.} & \textbf{ADE20k} & \textbf{Citysc.} & \textbf{VOC}  \\
        \midrule
         Qwen2-VL-ViT & 448 & 31.4 & 55.5 & 67.5 \\
        VersaViT & 448 & 42.3 \textcolor{darkgreen}{(+10.9)} & 66.7 \textcolor{darkgreen}{(+11.2)} & 81.0 \textcolor{darkgreen}{(+13.5)} \\
        \midrule
        Qwen2-VL-ViT & 560 & 33.6 & 57.6 & 67.5 \\
        VersaViT & 560 & 49.6 \textcolor{darkgreen}{(+16.0)} & 74.5 \textcolor{darkgreen}{(+16.9)} & 86.6 \textcolor{darkgreen}{(+19.1)} \\
        \bottomrule
    \end{tabular}
    % }
    \label{tab:supp_seg_resolution}
\end{table}

%% file: supp_tables/classification.tex
\begin{table}[htbp]
\footnotesize
\centering
\caption{\textbf{Performance of zero-shot image classification.}}
\vspace{-0.2cm}
\setlength\tabcolsep{6.2mm}
\begin{tabular}{lcc}
\toprule
\multirow{2}{*}{\textbf{Method}} & \multicolumn{1}{c}{\textbf{ImageNet}} & {\textbf{MMVP}} \\
\cmidrule(r){2-2} \cmidrule(l){3-3}
& Validation & Avg. \\
\midrule
% \midrule was missing after the header row
% Placeholder content added:
CLIP-L~\cite{radford2021learning} & 75.5 & 20.0 \\
EVA-CLIP-8B~\cite{sun2024eva} & 83.5 & - \\
SigLIP~\cite{zhai2023sigmoid} & 83.2 & 37.0 \\
SigLIP~2~\cite{tschannen2025siglip} & 84.1 & 34.8 \\
DINOv3~\cite{simeoni2025dinov3} & 82.3 & - \\
\midrule 
\multicolumn{3}{l}{\textbf{Aligned on 30M image-text pairs}} \\
Ours & 67.9 & 25.9 \\
Ours$^*$ & 75.6 & 32.6 \\
\bottomrule
\end{tabular}
\label{exp:classification}
\end{table}